%% file: main.tex
\documentclass[journal]{IEEEtran}

\usepackage{amsmath,amsfonts}
\usepackage{url}
\usepackage{booktabs}
\usepackage[ruled,vlined]{algorithm2e}
\usepackage{float}
\usepackage{siunitx}
\usepackage{adjustbox}
\usepackage[table]{xcolor}

\usepackage{hyperref}  
\usepackage{cite}      
\usepackage{multirow}

\usepackage[numbers,sort&compress]{natbib}

\definecolor{taskgray}{gray}{0.9}
\begin{document}

\title{Automating Parent Selection Configuration in \\ Genetic Programming with Agentic AI}

\author{Jose Guadalupe Hernandez\textsuperscript{\dag}, Jui-Hsuan Chang\textsuperscript{\dag}, Anil Kumar Saini\textsuperscript{\dag}, Xi Li\textsuperscript{\dag}, Jason H. Moore
\thanks{\textsuperscript{\dag}These authors contributed equally to this work.}
\thanks{Author Affiliations: Department of Computational Biomedicine, Cedars-Sinai Medical Center, West Hollywood, CA, 90069, USA.}}

\markboth{Journal of \LaTeX\ Class Files,~Vol.~14, No.~8, August~2021}%
{Shell \MakeLowercase{\textit{et al.}}: A Sample Article Using IEEEtran.cls for IEEE Journals}

\IEEEpubid{0000--0000/00\$00.00~\copyright~2021 IEEE}

\maketitle

\begin{abstract}
We investigate whether agentic artificial intelligence can automate parts of the process of designing genetic programming systems by introducing an agentic framework that identifies and implements parent selection algorithms using large language model (LLM) reasoning and retrieval-augmented generation.
Using symbolic regression as a test bed, we first conduct an ablation study across four LLM types to evaluate the effects of agentic reasoning and retrieval on generated algorithm categories, validity, implementation similarity, and downstream performance.
Results show that these components substantially influence the types of algorithms generated, but their downstream performance largely depends on the underlying LLM.
The strongest configuration, the full agentic setup with 5 mini (5 mini--AR), consistently generated established $\epsilon$-lexicase implementations while maintaining competitive downstream performance.
We then benchmark this configuration against fixed implementations of tournament selection and semi-dynamic MAD $\epsilon$-lexicase.
Across six symbolic regression problems, 5 mini--AR performed similarly to $\epsilon$-lexicase while generally outperforming tournament selection.
These findings demonstrate the potential of agentic AI to translate domain knowledge into generating executable components, providing a step toward automated configuration and design of evolutionary systems.
\end{abstract}

\begin{IEEEkeywords}
Agentic artificial intelligence, genetic programming, large language models, parent selection, retrieval-augmented generation, symbolic regression.
\end{IEEEkeywords}

\input{Text/introduction}
\input{Text/background}
\input{Text/agent}
\input{Text/experimental-setup}
\input{Text/results-discussion}
\input{Text/limitations}
\input{Text/conclusion}

\section*{Acknowledgments}
We thank the Department of Computational Biomedicine at Cedars-Sinai Medical Center for providing high-performance computing resources, and Alina Geiger for helpful comments and suggestions.
The work was supported by NIH grants P3O AG094848, U01 AG066833, and R01 LM014572 awarded to JHM.
 \bibliographystyle{IEEEtran}
 \bibliography{reference,software}

\newpage

 




\vfill

\end{document}

%% file: Text/introduction.tex
\section{Introduction}
\label{sec:intro}


\IEEEPARstart{S}{ymbolic} regression (SR) searches for mathematical expressions that model relationships between input variables and the corresponding outputs \cite{o2009riccardo}.
Genetic programming (GP) has demonstrated strong performance for SR relative to a range of non-GP methods \cite{la2021contemporary,makke2024review,dong2025symbolic}.
A key strength of GP is its flexibility, allowing representations, evaluation procedures, selection mechanisms, and variation operators to be tailored to a particular problem.
However, this flexibility introduces a large design space of interdependent algorithmic and parameter choices that can influence performance \cite{weise2012traps,crepinsek2013exp}.
Identifying effective configurations can therefore require substantial expertise and experimentation, posing a significant barrier to the broader adoption of GP.
Automating these design decisions could reduce this burden while enabling GP systems to adapt their components to specific problem domains.


Agentic artificial intelligence (AI) provides a potential approach for automating complex design tasks by extending large language models (LLMs) with capabilities for autonomous reasoning, planning, and interaction with external tools and information sources \cite{li2026aai,hosseini2025aai}.
Unlike standalone LLMs that generate responses directly from a prompt, AI agents can iteratively reason about a task, retrieve relevant information, execute actions, and use the resulting observations to guide subsequent decisions.
These capabilities are particularly relevant to GP system design, where selecting an appropriate component may require both domain knowledge and the ability to translate that knowledge into an executable implementation.


Saini et al. \cite{saini2026agp} proposed \textit{Agentic GP}, a conceptual framework for automatically constructing GP systems using agentic AI.
The framework employs a high-level composer agent to coordinate specialized agents responsible for four fundamental GP components: solution representation, parent selection, variation, and fitness evaluation.
Although Agentic GP provides a conceptual architecture for automating GP design, its feasibility depends on whether specialized agents can identify and implement effective GP components.
In this work, we investigate this question by isolating parent selection and evaluating the capabilities of a specialized agent responsible for its design.

\IEEEpubidadjcol


We propose an agentic framework for identifying and implementing parent selection algorithms for GP systems.
The framework combines an LLM with retrieval-augmented generation (RAG) using a curated corpus of parent selection literature, providing the agent with access to specialized knowledge on parent selection algorithms.
The framework is not specific to a particular GP problem domain; we use SR as a test bed to evaluate its ability to generate effective parent selection algorithms.
We evaluate the framework through two experiments.
First, an ablation study compares the complete agentic framework against a standalone LLM and an agent without RAG to determine how agentic reasoning and retrieval affect algorithm identification, implementation, and downstream performance.
Second, the strongest configuration identified by the ablation study is compared against established parent selection algorithms to determine whether automatically generated implementations can achieve competitive performance.

Our results demonstrate that agentic reasoning and retrieval substantially influence the parent selection algorithms generated, although their effects on algorithm validity and downstream performance depend on the underlying LLM.
Furthermore, the strongest configuration, 5 mini--AR, achieved performance comparable to a fixed $\epsilon$-lexicase implementation across all six SR problems while generally outperforming tournament selection, providing evidence that agentic AI can support the automated configuration of evolutionary systems.

%% file: Text/background.tex
\section{Background}
\label{sec:background}

\subsection{Large Language Models (LLMs)}


Large Language Models (LLMs) are probabilistic models trained on large-scale text corpora to generate sequences by predicting tokens conditioned on preceding context \cite{shen2023chatgpt,raiaan2024llm}.
Contemporary LLMs are predominantly based on the Transformer architecture \cite{vaswani2017attention} and are pretrained through self-supervised next-token prediction.
Through pretraining, LLMs acquire internal representations that capture linguistic patterns and broader domain knowledge, enabling them to perform a wide range of tasks using natural-language instructions and relevant context, without requiring task-specific architectures.


LLMs have increasingly been integrated into evolutionary computation (EC) systems \cite{wu2025ecllm,hemberg2025survey,morris2024llmge}.
Prior work has employed LLMs as variation operators for generating initial population and subsequent offspring \cite{lehman2024elm,huang2025llm,brownlee2024enhancing}, for repairing invalid solutions \cite{liventsev2023llm}, and for generating test cases to evaluate evolved solutions \cite{jorgensen2025policy,jorgensen2024llm}.
Other approaches have expanded the role of LLMs to multiple or all stages of the evolutionary cycle \cite{stein2025llamea,liu2024llmea}.
These studies demonstrate the potential for LLMs to support or automate components of evolutionary systems, motivating their use in automated GP design and construction.

\subsection{Retrieval-Augmented Generation (RAG)}


Retrieval-Augmented Generation (RAG) supplements the internal knowledge of an LLM with information retrieved from external knowledge sources \cite{lewis2020retrieval}.
Rather than relying solely on information encoded during pretraining, a RAG system retrieves information relevant to a query and provides it to the LLM as additional context for generating response.
This approach can improve response quality and reduce hallucinations by grounding generation in external information without requiring additional model training \cite{ayala2024reducing,salemi2024rag}.


A RAG pipeline generally consists of three stages: indexing, retrieval, and generation \cite{sharma2026rag}.
During indexing, documents are divided into chunks and encoded for retrieval.
Given a query, the retrieval stage identifies relevant chunks from the indexed corpus, which are then supplied as context during generation.
The effectiveness of RAG therefore depends on the quality of the knowledge corpus and retrieval process, as well as the LLM’s ability to interpret and use the retrieved information \cite{zhang2025hallucination,liu2024lost}.
Consequently, access to external information does not guarantee that an LLM will correctly apply it to the task.

\subsection{Agentic Large Language Models}


Agentic Large Language Models (LLMs) are a subset of agentic AI in which an LLM serves as the primary engine for reasoning, planning, and decision-making \cite{plaat2025agentic}.
Unlike conventional LLM applications that generate responses to individual prompts, agentic LLM systems embed the model within a control loop that enables iterative reasoning, action execution, progress monitoring, and interaction with external tools or environments.
This architecture enables LLMs to perform complex, multi-step tasks that require decisions to be revised based on intermediate observations.


In practice, an agentic LLM system maintains a state that records information accumulated throughout the task, such as previous interactions, actions, and outputs from external tools.
The control loop uses this state to coordinate successive reasoning and action cycles until a goal or termination condition is reached.
Combined with external tool use, this architecture enables iterative reasoning-and-acting protocols such as ReAct \cite{yao2023react}, in which the LLM selects actions, observes their outcomes, and uses those observations to inform subsequent decisions.


Agentic LLMs have increasingly been applied to the automated configuration and design of computational systems \cite{li2026aai,lu2026aisci}.
For example, AgentHPO \cite{liu2025agenthpo} and AutoML-GPT \cite{yun2023automl} use LLM-based agents for hyperparameter optimization, while AutoML-Agent \cite{trirat2025automlagent} uses multiple agents to automate stages of the machine learning pipeline.
Despite growing interest in agentic approaches to automated system design, their ability to identify and implement appropriate components of GP systems remains comparatively unexplored.

\subsection{Parent Selection}
\label{sub:background:parent_selection}






Tournament selection \cite{brindle1980genetic} is commonly used as a baseline for evaluating parent selection methods.
It selects a parent by randomly sampling a subset of the population and returning the best-performing individual from that subset.
Tournament size controls selection pressure \cite{goldberg1991comparitive,blickle1995comparison,back1996evolutionary}: small tournaments approach random selection, whereas tournaments approaching the population size increasingly favor the best individuals.


Lexicase selection \cite{helmuth2015lexicase} selects parents by considering performance on individual test cases rather than an aggregate fitness measure.
For each selection event, the test cases are randomly permuted and used sequentially to filter the candidate pool.
At each filtering step, only candidates whose performance matches the best performance on the current test case from the remaining candidates are retained.
Filtering continues until one candidate remains or all cases have been considered, after which a parent is randomly selected from the remaining candidates.
By allowing individuals that perform well on subsets of cases to be selected, lexicase selection preserves specialists and has demonstrated strong performance, particularly for program synthesis \cite{sobania2023survey}.


$\epsilon$-lexicase selection \cite{lacava2016epslex} relaxes the filtering criterion of lexicase by retaining candidates whose performance falls within an $\epsilon$ threshold of the best performance on each case.
Setting $\epsilon=0$ recovers standard lexicase selection, while larger values further relax the filtering criterion.
The choice of $\epsilon$ can substantially influence performance \cite{lacava2016epslex,hernandez2022expofexp,shakiba2024robustness}.
Rather than using a fixed value, $\epsilon$ can also be derived from the distribution of errors, commonly using the median absolute deviation (MAD) \cite{lacava2016epslex}.
Depending on how the best error and $\epsilon$ are computed, $\epsilon$-lexicase can be categorized into three variants \cite{lacava2019lex}:
\begin{itemize}
\item \textbf{Static}: The best error is computed from the population, while $\epsilon$ is computed once from population-level errors or specified as a fixed value.
\item \textbf{Semi-Dynamic}: The best error is recomputed from the current candidate pool, while $\epsilon$ is computed once from population-level errors or specified as a fixed value.
\item \textbf{Dynamic}: Both the best error and $\epsilon$ are recomputed from the current candidate pool.
\end{itemize}

\subsection{Symbolic Regression and Genetic Programming}


Symbolic regression has been a prominent application of GP since its introduction by Koza \cite{koza1994genetic}.
In GP-based SR, candidate mathematical expressions are commonly represented as expression trees and evolved using variation operators such as mutation and crossover.
Candidate fitness is determined by how well the resulting expression models the observed data, making evolutionary design choices, including parent selection, important to search performance.


Several studies have investigated the effect of parent selection on GP-based SR, with $\epsilon$-lexicase frequently outperforming standard lexicase and tournament selection \cite{geiger2023down,geiger2024comprehensive,geiger2025tournament}.
Geiger et al. \cite{geiger2023down}, for example, compared $\epsilon$-lexicase with other parent selection methods on six UCI regression problems and found that $\epsilon$-lexicase outperformed standard lexicase and tournament selection.
Their results also demonstrated that down-sampling can further improve the performance of $\epsilon$-lexicase.




Geiger et al. \cite{geiger2026performance} conducted a comprehensive comparison of parent selection methods for SR while controlling for other components of the GP system.
They evaluated six parent selection methods: fitness-proportionate selection, two tournament variants, and three $\epsilon$-lexicase variants.
Additionally, they combined each method with no down-sampling, Random Down-Sampling (RDS) \cite{hernandez2019dslex}, or Informed Down-Sampling (IDS) \cite{boldi2024informed}, yielding 18 configurations evaluated across 26 SR problems.
Their results showed that down-sampled configurations generally outperformed their non-down-sampled counterparts, with $\epsilon$-lexicase combined with RDS achieving the strongest overall performance.
Given the breadth of parent selection methods evaluated, we use this study to identify relevant parent selection literature for constructing the corpus (Sec. \ref{af:rag}) used by the proposed agentic framework.


More recently, Zhang et al.~\cite{zhang2026benchmarking} investigated zero-shot LLM generation of novel parent selection operators for GP-based SR.
They evaluated eight LLMs, generating ten independent operators per model and comparing their performance against automatic $\epsilon$-lexicase and tournament selection across twelve regression problems.
Their experimental setup is most closely related to the standalone LLM configuration evaluated in our work, which similarly uses zero-shot prompting; however, whereas Zhang et al. explicitly prompt LLMs to generate novel operators, we task the LLM with identifying and implementing an appropriate parent selection algorithm without requiring novelty.
We further extend this setting by evaluating agentic configurations both with and without retrieval of domain-specific literature and analyze not only downstream performance but also the algorithms generated and their implementations.

%% file: Text/agent.tex
\begin{figure*}[!ht]
\centering
\includegraphics[
    width=0.8\textwidth,
    trim={0 3.2cm 0 2.0cm},
    clip
]{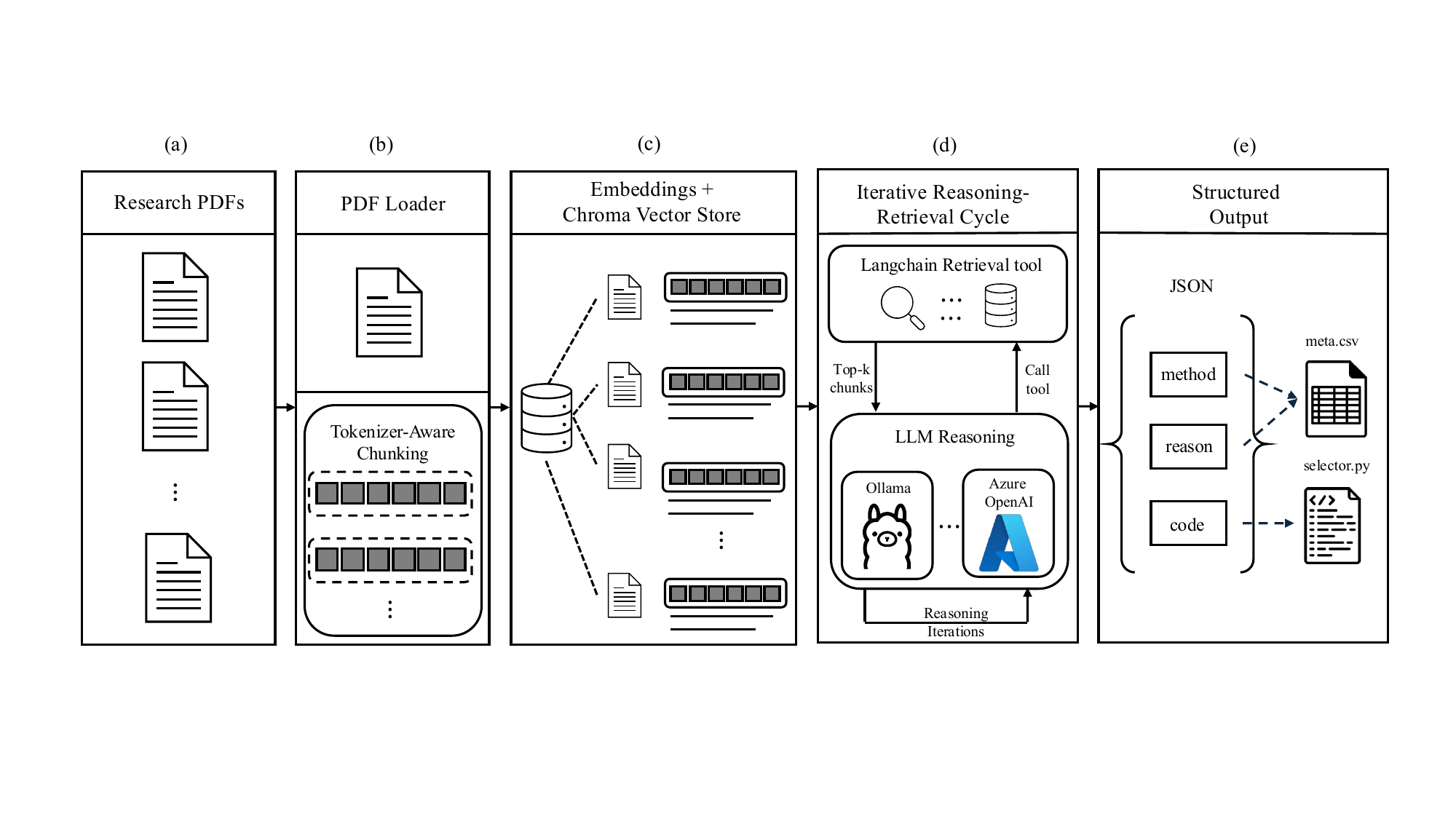}
\caption{
Overview of the proposed agentic framework for identifying and implementing parent selection algorithms:
(a) research literature is collected into a curated corpus,
(b) documents are loaded and partitioned into text chunks,
(c) chunks are embedded and stored in a vector database,
(d) the LLM-based agent iteratively reasons and may retrieve relevant literature to inform algorithm selection and implementation, and
(e) the selected method, rationale, generated code, and associated metadata are recorded for subsequent analysis.
}\label{fig:agentic_framework}
\end{figure*}

\section{Agentic Framework for Parent Selection}
\label{agentic-framework}


We propose an agentic framework for identifying and implementing parent selection algorithms for GP systems.
As illustrated in Fig.~\ref{fig:agentic_framework}, the framework combines an LLM-based agent with a RAG module containing domain-specific literature on parent selection.
The agent performs two primary tasks: (1) identifying an appropriate parent selection algorithm and (2) implementing the selected algorithm as a Python function.
We implement the framework using LangChain \cite{Chase_LangChain_2022}, which supports the iterative reasoning, state management, and tool invocation required by the agent.



The agent receives a system prompt defining its role as a GP specialist and a user prompt specifying the algorithm-identification and implementation tasks.
The user prompt also provides contextual information describing the GP system and the required implementation interface.
During algorithm identification, the agent autonomously determines whether to rely on its internal knowledge or invoke the RAG tool.
Retrieved information is incorporated into the agent state and made available during subsequent reasoning iterations, and the agent may invoke the RAG tool multiple times before selecting an algorithm.




After identifying a parent selection algorithm, the agent generates a Python function with the signature \texttt{selection(fitnesses)}, where \texttt{fitnesses} contains the population-level fitness matrix and the function returns the index of a selected parent.
The prompt provides only the information necessary to interpret this matrix and interface, without specifying or favoring a particular parent selection algorithm.
The final response is structured as JSON containing the selected \texttt{method}, a \texttt{reason} for its selection, and the generated \texttt{code}.
The framework extracts the generated code and records metadata, including generation latency, token counts, retrieval provenance, and parsing information, for subsequent analysis and reproducibility.
The complete prompts and implementation are provided in the Supplemental Material \footnote{\url{https://github.com/EpistasisLab/Agentic-Parent-Selection}}.

\subsection{Large Language Models (LLMs)}
\label{af:llms}


We evaluate four LLM model types: gpt-4o, gpt-4.1~mini, gpt-5~mini, and gpt-oss~20b, selected based on availability within our supported inference infrastructure.
The oss~20b model is deployed locally using Ollama, whereas the remaining models are accessed through Azure OpenAI.
Each model type is used to independently instantiate the agentic framework under the experimental configurations described in Section \ref{subsec:experiment1}.
To maintain consistent decoding settings where supported, we use a temperature of $0.3$ and nucleus sampling with $\mathrm{top}\text{-}p=0.8$.
GPT-5~mini does not expose these parameters and therefore uses its default decoding settings.

\subsection{RAG Module and Research Corpus}
\label{af:rag}

The RAG module is implemented as a callable LangChain tool that performs semantic similarity search over a vector database of scientific literature.
Research papers are partitioned using a token-based recursive text splitter with a chunk size of 256 tokens and 10\% overlap.
The resulting chunks are embedded using \texttt{mxbai-embed-large} and stored in a Chroma vector database, which is constructed offline and shared across agent invocations.
For each query, the tool retrieves the top-$k$ most similar chunks ($k=4$) and provides them to the agent as additional context.


The research corpus consists of nine papers selected based on the parent selection methods evaluated by Geiger et al. \cite{geiger2026performance} as a starting point.
For methods without down-sampling (NDS), we selected seminal or comprehensive papers describing the corresponding parent selection algorithms.
For Random Down-Sampling (RDS) and Informed Down-Sampling (IDS), we selected papers introducing or applying these sampling strategies to the considered parent selection methods.
Geiger et al. \cite{geiger2026performance} was used only to identify the parent selection methods represented in the corpus and was excluded as a retrieval source.

\input{Tables/corpus}


Table~\ref{tab:corpus} summarizes the nine papers and their coverage of the parent selection methods under NDS, RDS \cite{hernandez2019dslex}, and IDS  \cite{boldi2024informed}.
The corpus includes literature covering tournament \cite{blickle1995comparison}, fitness-proportionate \cite{blickle1995comparison}, lexicase \cite{helmuth2015lexicase}, $\epsilon$-lexicase \cite{lacava2019lex}, $\epsilon$-plexicase \cite{ding2023plex}, batch tournament \cite{vinicius2019batch}, and batch $\epsilon$-lexicase selection \cite{geiger2024comprehensive}.



%% file: Tables/corpus.tex
\begin{table}[h]
\scriptsize
\caption{Research papers included in the RAG corpus and their coverage of parent selection methods under no down-sampling (NDS), Random Down-Sampling (RDS), and Informed Down-Sampling (IDS).}
\label{tab:corpus}
\setlength{\tabcolsep}{5pt}
\renewcommand{\arraystretch}{1.35}
\begin{tabular}{lccc}
\toprule
\textbf{Parent Selection Method} & \textbf{NDS} & \textbf{RDS} & \textbf{IDS} \\
\midrule
Tournament & \cite{blickle1995comparison,geiger2023down,geiger2024comprehensive,helmuth2015lexicase,lacava2019lex,boldi2024informed,ding2023plex,vinicius2019batch} & \cite{boldi2024informed,geiger2024comprehensive} & \cite{boldi2024informed} \\
\hline
Fitness Proportionate      & \cite{blickle1995comparison,boldi2024informed} & \cite{boldi2024informed} & \cite{boldi2024informed} \\
\hline
Lexicase                   & \cite{geiger2023down,helmuth2015lexicase,lacava2019lex,boldi2024informed,hernandez2019dslex,ding2023plex} & \cite{boldi2024informed,ding2023plex,hernandez2019dslex} & \cite{boldi2024informed} \\
\hline
$\epsilon$-Lexicase         & \cite{geiger2023down,geiger2024comprehensive,lacava2019lex,ding2023plex,vinicius2019batch} & \cite{geiger2024comprehensive} & - \\
\hline
$\epsilon$-Plexicase       & \cite{ding2023plex,geiger2024comprehensive} & \cite{geiger2024comprehensive} & - \\
\hline
Batch Tournament           & \cite{geiger2024comprehensive,vinicius2019batch} & \cite{geiger2024comprehensive} & - \\
\hline
Batch $\epsilon$-Lexicase  & \cite{geiger2024comprehensive} & \cite{geiger2024comprehensive} & - \\
\bottomrule
\end{tabular}
\end{table}

%% file: Text/experimental-setup.tex
\section{Experimental Setup}
\label{sec:experimental_setup}

We evaluate the proposed framework through two experiments using a predefined GP system for SR.
Experiment 1 performs an ablation study across the four LLM model types described in Section~\ref{af:llms} to evaluate the effects of agentic reasoning and RAG on parent selection generation.
Experiment 2 compares the strongest configuration identified in Experiment 1 against representative parent selection algorithms observed during the ablation study.
Both experiments use the same GP system, benchmark problems, and evaluation procedure.

\subsection{Benchmark Problems and GP System}
\label{subsec:gp_framework}


We evaluate parent selection algorithms on six real-world SR problems from the UCI Machine Learning Repository \cite{uciml}, summarized in Table~\ref{tab:sr_tasks}.
For each problem, 70\% of the samples are allocated to training, with the remaining 30\% divided equally between validation and test sets.
Each configuration-problem pairing is evaluated using 21 independent GP replicates.

\input{Tables/problems}

The GP system is implemented using DEAP~\cite{fortin2012deap}, with individuals represented as expression trees.
The function set contains 14 operators: \texttt{add}, \texttt{sub}, \texttt{mul}, \texttt{div}, \texttt{sqrt}, \texttt{log}, \texttt{abs}, \texttt{neg}, \texttt{inv}, \texttt{max}, \texttt{min}, \texttt{sin}, \texttt{cos}, and \texttt{tan}, with protected implementations used where necessary.
The terminal set consists of the input features and an ephemeral random constant that samples uniformly from $[-1,1]$.
Initial populations are generated using ramped half-and-half initialization with tree depths ranging from 0 to 4.


Following Geiger et al.~\cite{geiger2026performance}, we use a population size of 500 and evolve the population for 50 generations.
Offspring are initially generated through crossover or mutation with probabilities of 0.8 and 0.2, respectively.
Offspring generated through crossover are subsequently mutated with probability 0.2.
Crossover is performed using one-point subtree crossover, while mutation randomly applies subtree replacement, node replacement, shrink, or insertion.
Tree depth is limited to 17 during variation, and individuals exceeding 100 nodes are considered invalid.
Individuals producing \texttt{NaN} or \texttt{Inf} predictions or otherwise failing evaluation are also considered invalid and removed before parent selection.
For each valid individual, squared errors are computed independently for each training sample and supplied to the parent selection procedure.
Selected parents are then used to generate the complete population of the next generation.
All GP components and parameters are held constant across experimental conditions, with only the parent selection procedure varying.


All successfully evaluated individuals encountered during evolution are retained in an archive.
After the final generation, duplicate tree structures are removed and all unique valid individuals are evaluated on the validation set.
The individual with the lowest validation mean squared error (MSE) is selected as the final model, with ties resolved uniformly at random.
The selected model is then evaluated on the held-out test set, and its test MSE is used to measure performance.

\subsection{Experiment 1: Ablation Study of Agentic Framework}
\label{subsec:experiment1}


We conduct an ablation study to determine the contributions of agentic reasoning and retrieval to parent selection algorithm generation.
For each of the four LLM types described in Section~\ref{af:llms}, we evaluate three configurations: the complete agentic framework with RAG (AR), the agentic framework without RAG (ANR), and the standalone LLM (LLM).
ANR preserves the agentic workflow while removing retrieval, whereas LLM removes both the agentic workflow and retrieval.
Prompts are modified accordingly so that information associated with an ablated component is excluded.


For each model--configuration--problem pairing, 21 parent selection algorithms are independently generated, with one algorithm used for each GP replicate.
Across the six SR problems, this yields 126 generated algorithms per model--configuration pairing and 1,512 algorithms overall.
We characterize the generated algorithms according to their execution success, classification success, algorithm family and implementation characteristics, code similarity, and performance.


Execution success indicates whether a generated parent selection implementation executes within the GP system without causing a runtime failure.
To assess classification success, we compare the parent selection method reported by the LLM or agent with a manual classification of its generated implementation.
Two GP experts (JGH and AKS) independently classify each implementation as tournament, lexicase, $\epsilon$-lexicase, or unknown, with disagreements resolved through discussion until consensus is reached.
Implementations classified as $\epsilon$-lexicase are further categorized as static, semi-dynamic, or dynamic according to the definitions in Section~\ref{sub:background:parent_selection}.
The method used to determine $\epsilon$ is also recorded and classified as MAD, a fixed numerical value, or Other, where Other denotes threshold calculations that do not correspond to these established approaches.
For descriptive analysis, fixed $\epsilon$ values are grouped into $\epsilon<0.1$ and $\epsilon\geq0.1$ to summarize the range of values generated without reporting each value separately.
For tournament selection, we additionally record the tournament size.
A classification is considered successful when the reported method agrees with the manually verified algorithm family and, for $\epsilon$-lexicase, its variant and threshold method.
MAD and fixed numerical values are considered valid threshold methods, whereas methods categorized as Other are considered classification failures.


Implementation-level similarity is quantified using the Winnowing document-fingerprinting algorithm \cite{schleimer2003winnowing}, implemented using the open-source \texttt{copydetect} package.\footnote{\url{https://github.com/blingenf/copydetect}}
For each model--configuration pairing, fingerprints are generated for all 126 implementations using a $5$-token $k$-gram size and a window size of 1.
Each implementation is compared against the other 125 implementations, with similarity measured as the proportion of its fingerprinted code that overlaps with each comparison implementation.
The median of these 125 similarity values is used as the program-level similarity score, yielding 126 similarity scores per model--configuration pairing.
Higher scores indicate greater implementation-level similarity within a model--configuration pairing.


Downstream performance is evaluated using the GP procedure described in Section~\ref{subsec:gp_framework}.
For non-executable implementations, we assign the highest test MSE observed among executable implementations for the corresponding SR problem and LLM model type so that performance comparisons capture failures of the complete algorithm-generation pipeline.


\subsection{Experiment 2: Benchmark Evaluation}
\label{subsec:experiment2}

Experiment 2 evaluates whether the highest-performing agentic configuration identified in Experiment 1 can generate parent selection algorithms that achieve competitive performance relative to fixed reference implementations.
Based on its combination of execution success, generation of established $\epsilon$-lexicase implementations, and performance in Experiment 1, we select 5 mini--AR for this evaluation.

We compare 5 mini--AR against two fixed parent selection algorithms: tournament selection (size 3) and semi-dynamic $\epsilon$-lexicase selection using MAD.
These algorithms are motivated by both the results of Experiment 1 and prior literature.
Tournament selection is commonly used as a baseline for evaluating parent selection methods, and tournament size 3 was the dominant tournament parameterization generated in Experiment 1.
Similarly, $\epsilon$-lexicase has demonstrated strong performance for SR, and for 5 mini, semi-dynamic $\epsilon$-lexicase and MAD-based thresholds were dominant across agent configurations.
Together, these reference algorithms provide standardized implementations representative of the parent selection behavior observed among the generated algorithms.

For 5 mini--AR, an independent parent selection implementation is generated for each of the 21 GP replicates of each problem; these implementations are also independent of the ones used in Experiment 1.
In contrast, the tournament and $\epsilon$-lexicase conditions use fixed, handcrafted implementations across all replicates.
All three conditions are evaluated using the same GP system, benchmark problems, replicate structure, and evaluation procedure described in Section~\ref{subsec:gp_framework}.
This experiment focuses on downstream performance to determine whether automatically generated parent selection implementations can perform competitively with fixed reference implementations.

\subsection{Statistical Analysis}
\label{subsec:statistics}

All statistical tests are two-sided with a significance level of $\alpha=0.05$.
GP replicates are independent across conditions, with each run using an independently generated GP seed and data split.
For all post hoc analyses, pairwise $p$-values within each significant omnibus comparison are adjusted using the Holm method to control the family-wise error rate.

For Experiment 1, execution and classification success are compared among agent configurations within each LLM type using the Fisher--Freeman--Halton (FFH) exact test, and when significant, followed by pairwise Fisher's exact tests (FET).
Code similarity is evaluated analogously using the Kruskal--Wallis (KW) test followed by pairwise Wilcoxon rank-sum (WRS) tests.
Downstream test MSE is evaluated separately for each SR problem using KW followed by pairwise WRS tests, comparing agent configurations within each LLM type in Experiment 1, and 5 mini--AR against the two fixed parent selection methods in Experiment 2.

%% file: Tables/problems.tex
\begin{table}[!h]
\scriptsize
\caption{Number of features and samples for all considered datasets.}
\label{tab:sr_tasks}
\renewcommand{\arraystretch}{1.0}
\setlength{\tabcolsep}{3pt}
\begin{tabular*}{\columnwidth}{@{\extracolsep{\fill}}lcc@{}}\toprule
\textbf{Task} & \textbf{Features} & \textbf{Samples} \\
\midrule
Airfoil        & 5  & 1503 \\
Concrete       & 8  & 1030 \\
Energy Cooling & 8  & 768  \\
Energy Heating & 8  & 768  \\
Housing        & 13 & 506  \\
Yacht          & 6  & 308  \\
\bottomrule
\end{tabular*}
\end{table}

%% file: Text/results-discussion.tex
\section{Results \& Discussion for Experiments 1 \& 2}
\label{sec:results-discussion}

\subsection{Experiment 1: Ablation Study}
\label{subsec:exp1}

\subsubsection{Parent Selection Algorithm Classification}


Table \ref{tab:algorithm_classification} reports the distribution of parent selection algorithm families assigned through manual classification of the algorithms generated by each LLM and agent configuration. 
Across all models, except 5 mini, both the LLM and ANR configurations generated tournament selection exclusively (100.0\%).
In contrast, incorporating the RAG module (AR configuration) shifted the dominant family from tournament to $\epsilon$-lexicase, although the magnitude of this shift varied across models.
For oss 20b, $\epsilon$-lexicase was the most frequently generated family, accounting for 44.4\% of the algorithms, while lexicase (28.6\%) and tournament (23.0\%) remained substantial. 
The 4o and 4.1 mini models generated $\epsilon$-lexicase in 76.2\% and 98.4\% of cases, respectively, indicating that RAG increased these models’ tendency to generate the intended family.


The 5 mini model exhibited a different pattern, generating $\epsilon$-lexicase as the dominant family across all agent configurations. 
In the LLM configuration, $\epsilon$-lexicase accounted for 85.7\% of the generated algorithms, indicating that the model frequently identified the intended parent selection algorithm without external retrieval. 
Under the ANR configuration, $\epsilon$-lexicase remained the dominant family, accounting for 73.0\% of all generated algorithms. 
However, tournament selection was chosen more frequently under ANR (25.4\%) than under the LLM configuration (12.7\%).
Incorporating the RAG module further strengthened this tendency, with the AR configuration generating $\epsilon$-lexicase exclusively (100.0\%), completely eliminating both tournament and lexicase selection outputs.

\input{Tables/family}


The results from the LLM and ANR configurations highlight differences in a models’ internal knowledge of parent selection.
For oss 20b, 4o, and 4.1 mini, the exclusive generation of tournament selection suggests that these models defaulted to the most familiar algorithm in the absence of external knowledge.
In contrast, the 5 mini model frequently generated $\epsilon$-lexicase without RAG, indicating that its pre-trained knowledge was sufficient to identify the intended parent selection family and suggesting a stronger internal representation of more recent parent selection methods.
Interestingly, the ANR configuration reduced the proportion of $\epsilon$-lexicase algorithms generated by the 5 mini model relative to both the LLM and AR configurations, suggesting that agentic reasoning may have led the model away from the intended algorithm.
Nevertheless, incorporating RAG consistently increased the generation of the target family across all evaluated models, demonstrating the value of external knowledge in guiding algorithm selection.

\input{Tables/lexicase-variant}

Table \ref{tab:eps-lex-variants} further categorizes the $\epsilon$-lexicase algorithms into three variants: static, semi-dynamic, or dynamic.
Across all models, static implementations were rare, accounting for no more than 10.9\% of the generated algorithms for any LLM–configuration pairing. 
Semi-dynamic and dynamic variants dominated, with their relative frequencies varying across LLMs. 
Under the AR configuration, oss 20b was the only model to favor the dynamic variant, which accounted for 69.6\% of its generated algorithms, whereas 4o and 4.1 mini predominantly generated semi-dynamic implementations (100.0\% and 86.3\%, respectively). 
In contrast, GPT-5 mini exhibited the greatest variation across agent configurations, with dynamic implementations accounting for 11.1\% under LLM, 28.3\% under ANR, and 38.9\% under AR.
These findings indicate that, beyond selecting an appropriate family, LLMs also differ in the specific implementation variants they produce. 
Furthermore, the prevalence of the semi-dynamic variant indicates that most model–configuration pairings favor computing the $\epsilon$ threshold using the entire population rather than the current candidate pool.

\subsubsection{Parent Selection Algorithm Parameterization}


By itself, correctly identifying the appropriate parent selection family and $\epsilon$-lexicase variant does not fully characterize the generated algorithms.
Table \ref{tab:eps-lex-params} examines the $\epsilon$-lexicase algorithms by detailing how the $\epsilon$ threshold is determined.
Implementations consistent with established approaches in the literature were classified as either MAD-based or fixed-$\epsilon$, with fixed values further separated into $\epsilon \geq 0.1$ and $\epsilon < 0.1$. 
We introduce the 0.1 cutoff as an analytical distinction because increasingly small $\epsilon$ values cause the selection behavior to approach that of standard lexicase. 
The Other category captures implementations that determine $\epsilon$ using approaches not identified among the established methods in the literature. 

\input{Tables/lex-params}

Differences in $\epsilon$ parameterization emerged across the evaluated LLMs. 
The oss 20b model exhibited the greatest departure from established approaches, with 83.9\% of its $\epsilon$-lexicase algorithms using methods categorized as Other and none using MAD.
In contrast, 4o predominantly generated fixed $\epsilon$ values, with 45.8\% using values greater than or equal to 0.1 and 41.7\% using values below 0.1. 
The 4.1 mini model produced a more diverse distribution, with fixed $\epsilon<0.1$ being the most common approach (46.0\%), followed by Other (25.8\%) and MAD-based implementations (25.0\%). 
The 5 mini model favored MAD-based computation across every agent configuration: 75.9\% under LLM, 68.4\% under ANR, and 89.7\% under AR. 
Notably, AR produced both the highest proportion of MAD-based implementations and the lowest proportion of Other implementations (10.3\%) for 5 mini, suggesting that retrieval shifted the model toward the established MAD-based approach while reducing the generation of parameterization strategies not identified in the literature.

\input{Tables/tournament-params}

The generated tournament selection algorithms exhibited strong model-specific preferences for tournament size (Table \ref{tab:tour-params}). The oss 20b model predominantly generated a tournament size of 3 across all agent configurations, accounting for 93.7\%, 96.0\%, and 62.1\% of tournament selection algorithms under LLM, ANR, and AR, respectively. 
However, AR produced greater variation for oss 20b, with 37.9\% of its tournament selection algorithms instead using a tournament size of 2. 
The 4o model also strongly favored a tournament size of 3, accounting for 100.0\% and 99.2\% of the tournament selection algorithms under LLM and ANR, respectively. 
In contrast, 4.1 mini consistently favored a tournament size of 7 (100.0\% under LLM and 98.4\% under ANR), as did 5 mini (100.0\% under LLM and 96.9\% under ANR). 
These results indicate that tournament size was largely dependent on the underlying LLM, with agent configuration generally introducing little variation in parameterization, aside from the increased diversity observed for oss 20b under AR.

\subsubsection{Algorithm Validity and Similarity}


Table \ref{tab:algorithm_analysis} summarizes algorithm validity and code similarity across LLM and agent configurations.
Validity is assessed by execution success, indicating error-free execution, and classification success, indicating agreement between the LLM/agent-assigned algorithm class and manual verification.
Code similarity quantifies token overlap among generated implementations, with each algorithm compared against the other generated algorithms to obtain similarity scores per LLM–configuration pairing.


Execution success was high across all LLM--configuration pairings.
All algorithms generated under the LLM and ANR configurations executed successfully.
Under the AR configuration, 4.1 mini and 5 mini achieved 100.0\% execution success, whereas oss 20b and 4o each successfully executed 123 of 126 generated algorithms (97.6\%).
The FFH test detected no significant differences in execution success among agent configurations for any model type.
Of the three execution failures observed for each of oss 20b and 4o, two in each were attributable to an IndexError. 
The remaining failure for oss 20b resulted from an infinite loop, whereas the remaining failure for 4o was due to a ValueError.
These results indicate that the agentic framework and retrieval components largely preserved the models’ ability to generate executable algorithms.

\input{Tables/proportions}


Classification success exhibited more variability across both agent configurations and model types. 
The FFH test detected significant differences among agent configurations for oss 20b, 4o, and 4.1 mini ($p<10^{-3}$).
For these models, the LLM and ANR configurations achieved 100.0\% classification success, whereas the AR configuration reduced success to 57.1\%, 88.9\%, and 61.1\% for oss 20b, 4o, and 4.1 mini, respectively. 
Pairwise FETs showed that AR differed significantly from both LLM and ANR for each of these models ($p<10^{-3}$).
In contrast, 5 mini's classification success was 77.0\% under LLM and 76.2\% under ANR, but increased to 85.7\% under AR. 
The AR configuration was associated with lower classification success for oss 20b, 4o, and 4.1 mini, whereas 5 mini achieved its highest classification success under AR.
Notably, 4o maintained the highest classification success across configurations, with success remaining at 88.9\% under AR despite the significant reduction relative to LLM and ANR.

Manual inspection indicated that these classification errors were primarily attributable to failures in identifying the correct $\epsilon$-lexicase variant.
In other words, the generated algorithms were often consistent with the overarching family (i.e.,`$\epsilon$-lexicase'), but the LLM/agent identified the incorrect variant.
These results highlight the importance of inspecting algorithms, particularly when distinctions among closely related variants are crucial.


The KW test identified significant differences in code similarity among agent configurations for all model types ($p<10^{-2}$).
For 4.1 mini, the LLM and ANR configurations both had median similarity scores of 1.000, although the LLM configuration exhibited substantially greater variability than ANR, with interquartile ranges of $[0.831, 1.000]$ and $[0.976, 1.000]$, respectively.
Pairwise WRS tests detected significant differences between all agent configurations ($p<10^{-2}$).
The lower similarity under AR likely reflects the shift to $\epsilon$-lexicase and variation in both the generated $\epsilon$-lexicase variants and the methods used to determine the $\epsilon$ threshold (Tables \ref{tab:eps-lex-variants} and \ref{tab:eps-lex-params}).
In contrast, the LLM and ANR configurations only generated tournament selection, with most implementations using a tournament size of 7 (Tables \ref{tab:algorithm_classification} and \ref{tab:tour-params}).
Despite generating the same algorithm family with similar parameterization, the wider similarity distribution under LLM indicates greater implementation variability than under ANR.


The 4o and oss 20b models exhibited similar patterns in code similarity, with pairwise WRS tests showing that AR differed significantly from both LLM and ANR ($p<10^{-3}$).
Both of these models exclusively generated tournament selection with tournament size 3 dominating under both LLM and ANR configurations (Tables \ref{tab:algorithm_classification} and \ref{tab:tour-params}). 
Despite this similarity in algorithm family and parameterization, oss 20b exhibited lower code similarity distributions than 4o, indicating greater implementation-level diversity.
In contrast, the AR configuration produced greater algorithm-family diversity generated by both oss 20b and 4o (Table \ref{tab:algorithm_classification}), which likely contributed to its significant differences in code similarity relative to both the LLM and ANR configurations.

\input{Tables/performance}


The 5 mini model exhibited relatively similar median code similarity scores across configurations $(0.857$–$0.874)$.
Despite these comparable medians, pairwise WRS tests showed that ANR differed significantly from both LLM and AR ($p<\alpha$).
This difference may reflect the distinct distribution of algorithms generated under ANR, which produced tournament selection more frequently (25.4\%) than LLM (12.7\%) or AR (0.0\%) (Table \ref{tab:algorithm_classification}).
Among the generated $\epsilon$-lexicase algorithms, ANR also exhibited the highest proportion of threshold calculations categorized as Other (29.3\%), compared with 19.4\% under LLM and 10.3\% under AR (Table \ref{tab:eps-lex-params}).
These differences suggest that the distinct code similarity distribution under ANR reflects differences in both the distribution of generated algorithm families and the implementation of $\epsilon$-lexicase algorithms that are not apparent from the median similarity scores alone.

\subsubsection{Parent Selection Algorithm Performance}


Table \ref{tab:model_results} reports the downstream performance of the generated parent selection algorithms across the six symbolic regression problems.
Performance is reported as the median test MSE [$Q_1$, $Q_3$] across replicates, with lower values indicating better performance.
Overall, the effect of agent configuration on downstream performance varied across model types: no significant differences were detected for oss 20b, AR degraded performance on several problems for 4o and 4.1 mini, and AR remained competitive for 5 mini.


For oss 20b, no significant differences among agent configurations were detected for any of the six problems.
The LLM and ANR configurations only generated tournament selection, whereas $\epsilon$-lexicase became the dominant family under AR, accounting for 44.4\% of the generated algorithms (Table \ref{tab:algorithm_classification}).
However, this shift toward $\epsilon$-lexicase did not translate into significant differences in downstream performance.
Inspection of the generated $\epsilon$-lexicase algorithms provides a possible explanation: 83.9\% used threshold calculations categorized as Other, while none used MAD (Table \ref{tab:eps-lex-params}).
Furthermore, 28.6\% of all algorithms generated under AR implemented lexicase selection (Table \ref{tab:algorithm_classification}), and an additional 16.1\% of the $\epsilon$-lexicase implementations used $\epsilon<0.1$ (Table \ref{tab:eps-lex-params}), resulting in behavior approaching lexicase.
Although retrieval altered the algorithms generated by oss 20b, it did not consistently produce established $\epsilon$-lexicase implementations or improve performance.
These findings suggest that oss 20b is poorly suited to the parent selection task considered in this study.


For 4o, the AR configuration exhibited higher MSE and greater performance variability than the LLM and ANR configurations.
The KW test detected significant differences for Airfoil, Energy Heating, and Yacht ($p<\alpha$).
Pairwise WRS tests showed that AR performed significantly worse than both LLM and ANR on Airfoil ($p<\alpha$) and Energy Heating ($p<10^{-2}$), and significantly worse than LLM on Yacht ($p<\alpha$).
This degradation coincided with a shift in the generated parent selection algorithms.
The LLM and ANR configurations exclusively generated tournament selection, whereas AR predominantly generated $\epsilon$-lexicase selection (76.2\%; Table \ref{tab:algorithm_classification}).
However, none of the 96 $\epsilon$-lexicase implementations generated under AR determined $\epsilon$ using MAD (Table \ref{tab:eps-lex-params}).
Instead, 44 implementations (45.8\%) used a fixed $\epsilon=0.1$, 40 (41.7\%) used $\epsilon<0.1$, and 12 (12.5\%) determined $\epsilon$ using methods categorized as Other.
These fixed-$\epsilon$ implementations differ substantially from the values identified by Geiger et al. \cite{geiger2023down}, who reported optimal values of $\epsilon=1.0$ for Concrete, Energy Heating, and Housing and $\epsilon=5.0$ for the remaining problems.
Although retrieval enabled 4o to generate the intended $\epsilon$-lexicase family more often, the resulting parameterizations differed from both MAD-based $\epsilon$-lexicase and previously identified effective fixed-$\epsilon$ settings, which may help explain the poor performance under AR.



The performance results for 4.1 mini followed a pattern similar to 4o.
The KW test detected significant differences for Energy Heating and Yacht ($p<10^{-2}$).
Pairwise WRS tests showed that AR performed significantly worse than both LLM and ANR on Energy Heating ($p<\alpha$) and Yacht ($p<10^{-2}$).
As with 4o, the LLM and ANR configurations only generated tournament selection, whereas AR generated $\epsilon$-lexicase in 98.4\% of cases (Table \ref{tab:algorithm_classification}).
Despite this near-exclusive generation of the intended algorithm family, only 25.0\% of the $\epsilon$-lexicase implementations determined $\epsilon$ using MAD (Table \ref{tab:eps-lex-params}).
Among the remaining implementations, 46.0\% used $\epsilon<0.1$, 25.8\% used methods categorized as Other, and 3.2\% used $\epsilon\geq0.1$.
Notably, only four of the 124 $\epsilon$-lexicase implementations generated under AR fell within the $\epsilon\geq0.1$ category.
Consequently, the increased frequency with which AR generated the intended $\epsilon$-lexicase family did not translate into improved performance.
Together with the 4o results, this finding further demonstrates that identifying the appropriate parent selection family alone is insufficient; the specific implementation and parameterization of $\epsilon$-lexicase substantially influence its effectiveness.

\begin{figure*}[!th]
\centering
\includegraphics[width=\textwidth]{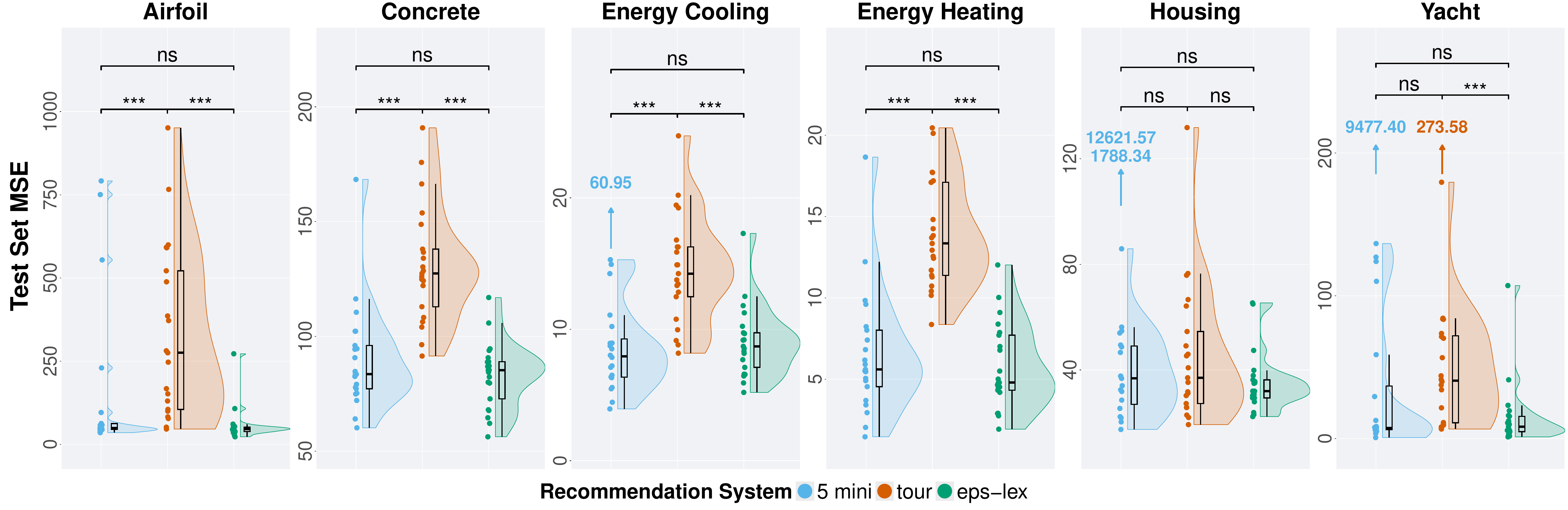}
\caption{Test-set MSE across the six SR tasks for 5 mini--AR, tournament selection (tour), and semi-dynamic MAD $\epsilon$-lexicase selection (eps-lex).
Raincloud plots show the distribution of the 21 independent replicates.
Brackets denote pairwise WRS comparisons: ns ($p\geq\alpha$), * ($p<\alpha$), and *** ($p<10^{-3}$).
For Energy Cooling, Housing, and Yacht, the y-axis is truncated for readability; excluded extreme values are indicated by arrows and remain included in the statistical analyses.}
\label{fig:benchmark_performance}
\end{figure*}


In contrast to the other model types, 5 mini exhibited competitive performance under AR.
AR achieved the lowest median test MSE on Airfoil, Energy Cooling, Energy Heating, and Yacht and produced performance nearly identical to LLM on Housing.
The KW test detected a significant difference only for Energy Cooling, where a pairwise WRS test showed that AR significantly outperformed LLM ($p<\alpha$).
Unlike the other models, 5 mini primarily generated $\epsilon$-lexicase even without retrieval, with $\epsilon$-lexicase accounting for 85.7\%, 73.0\%, and 100.0\% of the algorithms generated under LLM, ANR, and AR, respectively (Table \ref{tab:algorithm_classification}).
AR also produced the highest proportion of MAD-based $\epsilon$ thresholds (89.7\%; Table \ref{tab:eps-lex-params}) and the highest proportion of dynamic $\epsilon$-lexicase implementations (38.9\%; Table \ref{tab:eps-lex-variants}).
In comparison, the LLM and ANR configurations produced larger proportions of $\epsilon$-lexicase implementations using threshold calculations categorized as Other (19.4\% and 29.3\%, respectively) and generated tournament selection more frequently (12.7\% and 25.4\%, respectively).
Despite these differences in the distributions and implementations of the generated algorithms, downstream performance differed significantly on only one of the six problems.

For Energy Cooling, the significant difference between LLM and AR may be related to the specific algorithms generated under LLM. 
Among the LLM-generated algorithms evaluated on this problem, five $\epsilon$-lexicase implementations determined $\epsilon$ using methods categorized as Other, two used $\epsilon<0.1$, and three implemented tournament selection. 
Notably, five of the seven $\epsilon$-lexicase implementations using Other or $\epsilon<0.1$ threshold calculations produced test MSE values above the third quartile ($Q_3$) of the LLM performance distribution. 
This concentration among the poorer-performing implementations suggests that these $\epsilon$-lexicase parameterizations contributed to the lower performance of LLM relative to AR.

\subsection{Experiment 2: Benchmarking}
\label{sec:exp2-res}

Figure~\ref{fig:benchmark_performance} compares the performance of 5 mini--AR with fixed implementations of tournament selection and semi-dynamic MAD $\epsilon$-lexicase selection.
The KW test detected significant differences among parent selection methods for all problems except Housing.
Pairwise WRS tests found no significant differences between 5 mini--AR and $\epsilon$-lexicase on any of the six problems.
In contrast, tournament selection performed significantly worse than 5 mini--AR on Airfoil, Concrete, Energy Cooling, and Energy Heating ($p<10^{-3}$). 
Similarly, $\epsilon$-lexicase significantly outperformed tournament selection on these four problems and Yacht ($p<10^{-3}$).
Thus, 5 mini--AR generally reproduced the performance advantage of $\epsilon$-lexicase over tournament selection observed across these problems.

These results demonstrate that the agentic framework can generate parent selection implementations that achieve performance comparable to a fixed, expert-defined $\epsilon$-lexicase implementation.
Importantly, 5 mini--AR generates an independent implementation for each replicate, whereas the two reference methods use fixed implementations.
The absence of significant differences between 5 mini--AR and $\epsilon$-lexicase across all six problems suggests that the framework can repeatedly generate effective parent selection implementations despite implementation-level variation.
However, the extreme MSE values observed for individual 5 mini--AR replicates on Housing and Yacht indicate that generation is not uniformly reliable, motivating additional validation mechanisms before generated algorithms are incorporated into GP systems.

%% file: Tables/family.tex
\begin{table}[!h]
\scriptsize
\caption{Counts (percentages) of the parent selection algorithm families generated by each LLM and agent configuration. Each LLM–configuration pairing generated 126 algorithms. Boldface indicates the dominant family for each LLM–configuration pairing.}
\label{tab:algorithm_classification}
\renewcommand{\arraystretch}{1.0}
\setlength{\tabcolsep}{3pt}
\begin{tabular*}{\columnwidth}{@{\extracolsep{\fill}}c|cccccc@{}}\toprule
\multirow{2}{*}{} Model & Agent & $\epsilon$-Lexicase & Lexicase & Tournament & Unknown \\
Type & Config & Selection  & Selection & Selection & Selection   \\

\midrule

\multirow{3}{*}{\rotatebox[origin=c]{0}{oss 20b}}
& LLM & 0 (0.0\%) & 0 (0.0\%) & \textbf{126 (100.0\%)} & 0 (0.0\%) \\
& ANR & 0 (0.0\%) & 0 (0.0\%) & \textbf{126 (100.0\%)} & 0 (0.0\%) \\
& AR  & \textbf{56 (44.4\%)} & 36 (28.6\%) & 29 (23.0\%) & 5 (4.0\%) \\
\midrule

\multirow{3}{*}{\rotatebox[origin=c]{0}{4o}}
& LLM & 0 (0.0\%) & 0 (0.0\%) & \textbf{126 (100.0\%)} & 0 (0.0\%) \\
& ANR & 0 (0.0\%) & 0 (0.0\%) & \textbf{126 (100.0\%)} & 0 (0.0\%) \\
& AR  & \textbf{96 (76.2\%)} & 28 (22.2\%) & 0 (0.0\%) & 2 (1.6\%) \\
\midrule

\multirow{3}{*}{\rotatebox[origin=c]{0}{4.1 mini}}
& LLM & 0 (0.0\%) & 0 (0.0\%) & \textbf{126 (100.0\%)} & 0 (0.0\%) \\
& ANR & 0 (0.0\%) & 0 (0.0\%) & \textbf{126 (100.0\%)} & 0 (0.0\%) \\
& AR  & \textbf{124 (98.4\%)} & 0 (0.0\%) & 0 (0.0\%) & 2 (1.6\%) \\
\midrule

\multirow{3}{*}{\rotatebox[origin=c]{0}{5 mini}}
& LLM & \textbf{108 (85.7\%)} & 2 (1.6\%) & 16 (12.7\%) & 0 (0.0\%) \\
& ANR & \textbf{92 (73.0\%)} & 2 (1.6\%) & 32 (25.4\%) & 0 (0.0\%) \\
& AR  & \textbf{126 (100.0\%)} & 0 (0.0\%) & 0 (0.0\%) & 0 (0.0\%) \\
\bottomrule
\end{tabular*}
\end{table}

%% file: Tables/lexicase-variant.tex
\begin{table}[!h]
\scriptsize

\caption{Counts (percentages) of the $\epsilon$-lexicase implementation variants among the algorithms classified as $\epsilon$-lexicase in Table \ref{tab:algorithm_classification}. Percentages are computed relative to the total number of $\epsilon$-lexicase algorithms generated for each LLM–configuration pairing. Only pairings that generated at least one $\epsilon$-lexicase algorithm are included.}
\label{tab:eps-lex-variants}
\renewcommand{\arraystretch}{1.0}
\setlength{\tabcolsep}{3pt}
\begin{tabular*}{\columnwidth}{@{\extracolsep{\fill}}c|cccccc@{}}\toprule
\multirow{2}{*}{} Model & Agent & $\epsilon$-Lexicase & Static & Semi-Dynamic & Dynamic \\
Type & Config & Total  & Count & Count & Count   \\

\midrule

\multirow{1}{*}{\rotatebox[origin=c]{0}{oss 20b}}
& AR  & 56 & 0 (0.0\%) & 17 (30.4\%) & \textbf{39 (69.6\%)} \\
\midrule

\multirow{1}{*}{\rotatebox[origin=c]{0}{4o}}
& AR  & 96 & 0 (0.0\%) & \textbf{96 (100.0\%)} & 0 (0.0\%) \\
\midrule

\multirow{1}{*}{\rotatebox[origin=c]{0}{4.1 mini}}
& AR  & 124 & 0 (0.0\%) & \textbf{107 (86.3\%)} & 17 (13.7\%) \\
\midrule

\multirow{3}{*}{\rotatebox[origin=c]{0}{5 mini}}
& LLM & 108 & 5 (4.6\%) & \textbf{91 (84.3\%)} & 12 (11.1\%) \\
& ANR & 92 & 10 (10.9\%) & \textbf{56 (60.9\%)} & 26 (28.3\%) \\
& AR  & 126 & 3 (2.4\%) & \textbf{74 (58.7\%)} & 49 (38.9\%)  \\
\bottomrule
\end{tabular*}
\end{table}

%% file: Tables/lex-params.tex
\begin{table}[!h]
\scriptsize

\caption{Counts (percentages) of the methods used to compute the $\epsilon$ threshold for the generated $\epsilon$-lexicase algorithms in Table \ref{tab:algorithm_classification}. Percentages are relative to the total number of $\epsilon$-lexicase algorithms generated for each LLM–configuration pairing. Only pairings that generated at least one $\epsilon$-lexicase algorithm are included.}
\label{tab:eps-lex-params}
\renewcommand{\arraystretch}{1.0}
\setlength{\tabcolsep}{1.0pt}
\begin{tabular*}{\columnwidth}{@{\extracolsep{\fill}}c|cccccc@{}}\toprule
\multirow{2}{*}{} Model & Agent & $\epsilon$-lex & MAD & $\epsilon\geq0.1$  & $\epsilon < 0.1$ & Other\\
Type & Config & Total  & Count & Count & Count & Count \\

\midrule

\multirow{1}{*}{\rotatebox[origin=c]{0}{oss 20b}}
& AR  & 56 & 0 (0.0\%) & 0 (0.0\%) & 9 (16.1\%) & \textbf{47 (83.9\%)} \\
\midrule

\multirow{1}{*}{\rotatebox[origin=c]{0}{4o}}
& AR  & 96 & 0 (0.0\%) & \textbf{44 (45.8\%)} & 40 (41.7\%) & 12 (12.5\%) \\
\midrule

\multirow{1}{*}{\rotatebox[origin=c]{0}{4.1 mini}}
& AR  & 124 & 31 (25.0\%) & 4 (3.2\%) & \textbf{57 (46.0\%)} & 32 (25.8\%) \\
\midrule

\multirow{3}{*}{\rotatebox[origin=c]{0}{5 mini}}
& LLM & 108 & \textbf{82 (75.9\%)} & 0 (0.0\%) & 5 (4.6\%) & 21 (19.4\%) \\
& ANR & 92 & \textbf{63 (68.4\%)} & 0 (0.0\%) & 2 (2.2\%) & 27 (29.3\%) \\
& AR  & 126 & \textbf{113 (89.7\%)} & 0 (0.0\%) & 0 (0.0\%) & 13 (10.3\%) \\
\bottomrule
\end{tabular*}
\end{table}

%% file: Tables/tournament-params.tex
\begin{table}[!h]
\scriptsize
\caption{Counts (percentages) of tournament sizes among the algorithms classified as tournament selection in Table \ref{tab:algorithm_classification}. Percentages are relative to the total number of tournament selection algorithms generated for each LLM–configuration pairing. Only pairings that generated at least one tournament selection algorithm are included.}
\label{tab:tour-params}
\renewcommand{\arraystretch}{1.0}
\setlength{\tabcolsep}{1pt}
\begin{tabular*}{\columnwidth}{@{\extracolsep{\fill}}c|ccccccc@{}}

\toprule

Model & Agent & Tour. & \multicolumn{5}{c}{Tournament Size} \\

\cmidrule(lr){4-8}

Type & Config & Total & 2 & 3 & 5 & 7 & Top-2\% \\

\midrule

\multirow{3}{*}{\rotatebox[origin=c]{0}{oss 20b}}
& LLM & 126 & 8 (6.3\%) & \textbf{118 (93.7\%)} & 0 (0.0\%) & 0 (0.0\%) & 0 (0.0\%) \\
& ANR & 126 & 5 (4.0\%) & \textbf{121 (96.0\%)} & 0 (0.0\%) & 0 (0.0\%) & 0 (0.0\%) \\
& AR  & 29  & 11 (37.9\%) & \textbf{18 (62.1\%)} & 0 (0.0\%)  & 0 (0.0\%) & 0 (0.0\%) \\
\midrule

\multirow{2}{*}{\rotatebox[origin=c]{0}{4o}}
& LLM & 126 & 0 (0.0\%) & \textbf{126 (100.0\%)} & 0 (0.0\%) & 0 (0.0\%) & 0 (0.0\%) \\
& ANR & 126 & 0 (0.0\%) & \textbf{125 (99.2\%)} & 1 (0.8\%) & 0 (0.0\%) & 0 (0.0\%) \\
\midrule

\multirow{2}{*}{\rotatebox[origin=c]{0}{4.1 mini}}
& LLM & 126 & 0 (0.0\%) & 0 (0.0\%) & 0 (0.0\%) & \textbf{126 (100.0\%)} & 0 (0.0\%) \\
& ANR & 126 & 0 (0.0\%) & 1 (0.8\%) & 1 (0.8\%) & \textbf{124 (98.4\%)} & 0 (0.0\%) \\
\midrule

\multirow{2}{*}{\rotatebox[origin=c]{0}{5 mini}}
& LLM & 16 & 0 (0.0\%) & 0 (0.0\%) & 0 (0.0\%) & \textbf{16 (100.0\%)} & 0 (0.0\%) \\
& ANR & 32 & 0 (0.0\%) & 0 (0.0\%) & 0 (0.0\%) & \textbf{31 (96.9\%)} & 1 (3.1\%) \\
\bottomrule
\end{tabular*}
\end{table}

%% file: Tables/proportions.tex
\begin{table}[h]
\scriptsize
\caption{
Execution success, classification success, and code similarity across LLMs and agent configurations. 
Success metrics are reported as counts (percentages), and code similarity as median [$Q_1$, $Q_3$]. 
A dagger ($\dagger$) indicates a significant difference from other agent configurations within the same model.
}
\label{tab:algorithm_analysis}
\renewcommand{\arraystretch}{1.1}
\setlength{\tabcolsep}{3pt}
\begin{tabular*}{\columnwidth}{@{\extracolsep{\fill}}c|ccccc@{}}\toprule
\multirow{2}{*}{} Model & Agent & Execution & Classification & Code \\
Type & Config & Success  & Success & Similarity   \\
\midrule

\multirow{3}{*}{\rotatebox[origin=c]{0}{oss 20b}}
& LLM & 126 (100.0\%) & 126 (100.0\%)  & 0.872 [0.818, 0.904] \\
& ANR & 126 (100.0\%) & 126 (100.0\%)  & 0.890 [0.824, 0.920] \\
& AR  & 123 (97.6\%)  & 72 (57.1\%)$^\dagger$  & 0.829 [0.765, 0.881]$^\dagger$ \\
\midrule

\multirow{3}{*}{\rotatebox[origin=c]{0}{4o}}
& LLM & 126 (100.0\%) & 126 (100.0\%) & 1.000 [0.971, 1.000] \\
& ANR & 126 (100.0\%) & 126 (100.0\%) & 1.000 [0.971, 1.000] \\
& AR  & 123 (97.6\%)  & 112 (88.9\%)$^\dagger$ & 0.948 [0.918, 0.960]$^\dagger$ \\
\midrule

\multirow{3}{*}{\rotatebox[origin=c]{0}{4.1 mini}}
& LLM & 126 (100.0\%) & 126 (100.0\%)   & 1.000 [0.831, 1.000]$^\dagger$ \\
& ANR & 126 (100.0\%) & 126 (100.0\%)   & 1.000 [0.976, 1.000]$^\dagger$ \\
& AR  & 126 (100.0\%) &  77 (61.1\%)$^\dagger$ & 0.828 [0.767, 0.885]$^\dagger$ \\
\midrule

\multirow{3}{*}{\rotatebox[origin=c]{0}{5 mini}}
& LLM & 126 (100.0\%)  & 97 (77.0\%) & 0.874 [0.830, 0.907] \\
& ANR & 126 (100.0\%)  & 96 (76.2\%) & 0.857 [0.776, 0.902]$^\dagger$ \\
& AR  & 126 (100.0\%)  & 108 (85.7\%) & 0.864 [0.823, 0.917] \\
\bottomrule
\end{tabular*}
\end{table}

%% file: Tables/performance.tex
\begin{table*}[!ht]
\scriptsize
\caption{Median test-set MSE [$Q_1$, $Q_3$] across replicates for each problem, model type, and agent configuration; lower values indicate better performance. 
Values are rounded to two decimal places. 
Configurations are labeled $^{a}$LLM, $^{b}$ANR, and $^{c}$AR; superscripts preceding a result indicate configurations from which that result differs significantly according to Holm-adjusted pairwise Wilcoxon rank-sum tests following a significant Kruskal–Wallis omnibus test.}
\label{tab:model_results}
\renewcommand{\arraystretch}{1.1}
\setlength{\tabcolsep}{4.5pt}
\begin{tabular}{cccccccc}
\toprule
\multirow{2}{*}{    } & Model \& & Airfoil & Concrete & Energy Cooling & Energy Heating & Housing & Yacht \\
& Config & (air)  & (con)  & (enc) & (enh) & (hse)  & (yat)  \\
\midrule

\multirow{3}{*}{\rotatebox[origin=c]{90}{oss 20b}}
& $^{a}$LLM & 604.27 [137.34, 720.23] & 142.27 [130.29, 161.05] & 15.08 [13.48, 17.81] & 15.11 [11.94, 17.04] & 52.89 [32.53, 64.01] & 50.13 [20.82, 57.71] \\
& $^{b}$ANR & 189.02 [88.82, 417.27] & 131.54 [108.05, 161.69] & 13.24 [11.46, 15.86] & 16.45 [12.46, 18.73] & 51.95 [34.57, 58.63] & 30.46 [21.04, 52.86] \\
& $^{c}$AR  & 243.53 [106.06, 783.93] & 149.67 [128.56, 199.53] & 13.83 [9.89, 16.96]  & 16.88 [12.01, 22.57] & 58.65 [35.70, 75.27] & 43.71 [14.58, 126.65] \\
\midrule

\multirow{3}{*}{\rotatebox[origin=c]{90}{4o}}
& $^{a}$LLM & $^{c}$217.61 [110.28, 419.51]     & 119.57 [102.84, 141.80] & 14.65 [13.09, 16.42] & $^{c}$14.96 [12.76, 16.29]   & 46.48 [38.90, 57.90] & $^{c}$27.34 [21.08, 31.54] \\
& $^{b}$ANR & $^{c}$244.28 [176.07, 593.85]     & 145.65 [119.10, 171.28] & 15.37 [13.42, 17.30] & $^{c}$15.53 [11.82, 19.24]   & 50.68 [42.78, 57.30] & 33.93 [20.13, 45.81]  \\
& $^{c}$AR  & $^{ab}$1084.86 [229.01, 1488.83] & 208.70 [107.84, 296.71]  & 16.80  [14.13, 56.31] & $^{ab}$51.68 [16.52, 67.63]  & 69.61 [34.42, 85.61] & $^{a}$104.15 [26.30, 135.55] \\
\midrule

\multirow{3}{*}{\rotatebox[origin=c]{90}{4.1 mini}}
& $^{a}$LLM   & 78.50 [42.77, 217.86] & 94.66 [89.33, 111.01]  & 11.10 [8.01, 14.18] & $^{c}$11.88 [8.21, 13.95]  & 35.95 [30.53, 45.64] & $^{c}$7.16 [6.09, 16.25] \\
& $^{b}$ANR  & 52.36 [46.07, 119.41] & 96.18 [84.81, 115.80]  & 10.74 [9.57, 12.30] & $^{c}$9.72 [6.35, 11.59]   & 38.13 [26.04, 66.48] & $^{c}$10.89 [4.79, 19.28] \\
& $^{c}$AR  & 90.30 [59.42, 773.89] & 109.87 [87.42, 143.61] & 13.62 [8.53, 17.04] & $^{ab}$15.27 [13.22, 32.64] & 48.30 [33.85, 67.46] & $^{ab}$47.05 [12.52, 106.20] \\
\midrule

\multirow{3}{*}{\rotatebox[origin=c]{90}{5 mini}}
& $^{a}$LLM   & 49.89 [45.71, 126.57] & 85.76 [72.54, 120.01] & $^{c}$10.59 [7.74, 16.21] & 8.44 [5.15, 9.69] & 28.35 [26.20, 37.58] & 8.85 [3.89, 19.26] \\
& $^{b}$ANR  & 83.88 [43.64, 309.66] & 93.14 [89.75, 114.14] & 8.13 [6.87, 10.28] & 6.87 [4.72, 11.73] & 35.29 [27.21, 43.30] & 14.67 [7.86, 25.99] \\
& $^{c}$AR  & 46.35 [38.92, 53.03]  & 97.90 [72.35, 107.52] & $^{a}$7.02 [5.89, 8.96]  & 5.85 [2.71, 8.36]  & 28.40 [19.56, 36.76] & 8.71 [3.32, 24.27] \\
\bottomrule
\end{tabular}
\end{table*}

%% file: Text/limitations.tex
\section{Limitations and Future Work}

This study evaluates the proposed agentic framework for generating a single GP component for a particular problem domain.
We isolate parent selection to determine whether a specialized agent can identify and implement effective algorithms for a predefined GP system.
Although the framework is designed to operate independently of a specific problem domain, its evaluation is limited to six SR problems.
Future work should evaluate the framework across additional GP domains and extend it to other components (e.g., representation, variation, and fitness evaluation).
Integrating these specialized agents within a larger system would provide a more complete evaluation of the Agentic GP vision.

The effectiveness of the framework depends on the underlying LLM and the external knowledge available through RAG.
Experiment 1 highlighted differences among model types in the algorithms generated and their downstream performance, indicating that the choice of LLM can influence the reliability of automated GP design.
Similarly, the RAG module relies on a small, expert-curated corpus and a fixed retrieval configuration.
Future work should evaluate additional LLMs and investigate how corpus composition and retrieval strategies affect algorithm generation.
Such analyses could help identify the model and retrieval characteristics most important for reliably translating domain knowledge into effective algorithms.

Finally, the framework generates a parent selection algorithm without validating its effectiveness or using its downstream performance to revise the generated implementation.
Although 5 mini--AR achieved performance comparable to the fixed $\epsilon$-lexicase implementation across Experiment 2, individual generated implementations produced extreme performance failures on some replicates (Figure \ref{fig:benchmark_performance}).
Future work should incorporate automated validation mechanisms that evaluate generated implementations before their deployment and enable the agent to revise unsuccessful algorithms.
More broadly, incorporating GP performance as feedback would enable an iterative generate--evaluate--revise process in which the agent adapts its decisions based on empirical observations, moving toward closed-loop automated GP design.

%% file: Text/conclusion.tex
\section{Conclusion}
\label{sec:conclusion}

This work expands the toolkit for the automated design of evolutionary systems by introducing an agentic framework that uses LLM-based reasoning and RAG to identify and implement parent selection algorithms for GP.
Through an ablation study, we demonstrate that the effectiveness of this approach depends on the underlying LLM and agent configuration, with RAG module substantially changing the generated algorithms but not uniformly improving their validity or downstream performance.
Among the evaluated configurations, the agentic setup for 5 mini with RAG enabled (5 mini--AR) reliably generated established $\epsilon$-lexicase implementations and achieved competitive performance, motivating its selection for further evaluation.
When benchmarked against fixed implementations of tournament selection and semi-dynamic MAD $\epsilon$-lexicase, 5 mini--AR performed similarly to $\epsilon$-lexicase across all six SR problems and generally outperformed tournament selection.
These findings demonstrate that agentic LLMs can move beyond assisting practitioners with isolated tasks toward autonomously translating domain knowledge into generating effective algorithmic components.
More broadly, this work provides initial evidence that agentic AI can serve as a mechanism for automating configuration decisions, providing a foundation for future systems that autonomously design, evaluate, and refine multiple interacting components of an evolutionary algorithm.